\documentclass[letterpaper, 10 pt, conference, english]{ieeeconf}  

\IEEEoverridecommandlockouts                              
\usepackage[T1]{fontenc}
\usepackage[latin9]{inputenc}
\usepackage{verbatim}
\usepackage{amsmath}
\usepackage{graphicx}

\usepackage{amssymb}
\usepackage{algpseudocode}
\usepackage{algorithm}
\usepackage{mathrsfs}

\makeatletter

\title{\LARGE \bf
Network Anomaly Detection: A Survey and Comparative Analysis of Stochastic and Deterministic Methods\authorrefmark{1}}

\author{Jing~Wang, \authorrefmark{2} 
\thanks{* Research partially supported by the ARO under grants
W911NF-11-1-0227 and 61789-MA-MUR, by the 
NSF under grants EFRI-0735974, 
CNS-1239021 and IIS-1237022, by the AFOSR under grant FA9550-12-1-0113,
and by the ONR under grants N00014-09-1-1051 and N00014-10-1-0952.}
\thanks{$\dagger$ Division of Systems Eng., Boston University, email:
  {\tt wangjing@bu.edu}.}
Daniel~Rossell,\authorrefmark{3} \thanks{$\ddagger$ Department of Electrical and
Computer Eng., Boston University, email: {\tt aggie0642@googlemail.com}.}
Christos~G.~Cassandras,\authorrefmark{4} and Ioannis~Ch.~Paschalidis
\authorrefmark{4} 
\thanks{$\S$ Department of Electrical and
Computer Eng., and Division of Systems Eng., Boston
University, 8 Saint Mary's St., Boston, MA 02215, e-mails: {\tt
  cgc@bu.edu} and {\tt yannisp@bu.edu}.}
}

\usepackage{babel}
\usepackage{url}

\def\msC{{\mathcal C}}
\def\msT{{\mathcal T}}
\def\msI{{\mathcal I}}
\def\1{{\mathbf 1}}
\def\c{{\mathbf c}}
\def\p{{\mathbf p}}
\def\q{{\mathbf q}}

\def\msS{{\mathcal S}}
\def\s{{\mathbf s}}
\def\x{{\mathbf x}}
\def\z{{\mathbf z}}
\def\f{{\mathbf f}}
\def\msF{{\mathcal F}}
\def\msX{{\mathcal X}}
\def\msY{{\mathcal Y}}
\def\bsigma{{\boldsymbol \sigma}}
\def\brho{{\boldsymbol \rho}}
\def\msD{{\mathcal D}}

\def\msG{{\mathcal G}}
\def\bmu{{\boldsymbol \mu}}
\def\msE{{\mathcal E}}
\def\bmsE{{\boldsymbol {{\mathcal E}}}}
\def\bnu{{\boldsymbol \nu}}

\def\g{{\mathbf g}}
\def\Q{{\mathbf Q}}
\def\bPi{{\boldsymbol \Pi}}

\def\u{{\mathbf u}}
\def\v{{\mathbf v}}
\def\s{{\mathbf s}}
\def\msZ{{\mathcal Z}}
\def\z{{\mathbf z}}

\begin{document}
\maketitle

\begin{abstract}
  We present five methods to the problem of network anomaly
  detection. These methods cover most of the common techniques in the
  anomaly detection field, including Statistical Hypothesis Tests (SHT),
  Support Vector Machines (SVM) and clustering analysis.  We evaluate
  all methods in a simulated network that consists of nominal data,
  three flow-level anomalies and one packet-level attack. Through
  analyzing the results, we point out the advantages and disadvantages
  of each method and conclude that combining the results of the
  individual methods can yield improved anomaly detection results.
\end{abstract}


\section{Introduction}
\label{sec:intro}
A network anomaly is any potentially malicious traffic that has implications
for the security of the network. It is of particular importance to the
prevention of zero-day attacks, i.e., attacks not previously seen, and malicious
data exfiltration. These are key areas of concern for both government and
corporate entities.

From the perspective of methodology, network anomaly detection methods can be
classified as stochastic and deterministic. Stochastic methods fit reference
data to a probabilistic model and evaluate the fitness of the new traffic with
respect to this model~\cite{hareesh2011anomaly, lakhina2005mining, Lee2001,
Perona2010, Zhang2007}. The evaluation can be done using Statistical Hypothesis
Testing (SHT)~\cite{Frakt1998,lehmann2005testing, LockeWangPasTech,
Paschalidis2009}.
Deterministic methods, on the other hand, try to partition the feature space
into ``normal'' and ``abnormal'' regions through a deterministic decision
boundary. The boundary can be determined using methods like Support Vector Machine (SVM),
particularly 1-class SVM~\cite{hastie2001elements, Perdisci2006,
shon2007hybrid}, and clustering analysis~\cite{anderberg1973cluster,
gu2008botminer}.

From the perspective of data, network anomaly methods can be either
packet-based~\cite{hareesh2011anomaly,Mahoney2001},
flow-based~\cite{Androulidakis2008, Manikopoulo2006} or
window-based~\cite{Lee2001}.  Packet-based methods evaluate the raw packets
directly while both flow-based and window-based methods aggregate the packets
first. Flow-based methods evalulate each flow individually, which is defined as a
collection of packets with similar properties. Flows are considered as a good
tradeoff between cost of collection and level of detail~\cite{Sommers2011}.
Window-based methods group consecutive packets or flows based on a sliding
window.

The main goal of this paper is to discuss the advantages and distadvantages of
each category of methods for different applications.  This paper presents five
methods, covering most of the categories discussed above, to the problem of
host-based network anomaly detection and provides a comparative analysis. The
first four methods are revisions of authors' previous work with collaborators
and the last method is new. 

The first two methods are based on SHT, utilizing results from Large Deviations
Theory (LDT)~\cite{dembo2009large} to compare current network traffic to
probability laws governing nominal network traffic.  The two methods fit
traffic, which is a sequence of flows, with probabilistic models under i.i.d.\
and Markovian assumptions, respectively.  We refer to these two methods as
\emph{model-free} and \emph{model-based} methods.

The next two methods are based on an 1-class SVM\@. In the first of the two
methods, individual data transmissions from a given source are examined
independently from neighboring transmissions, producing a flow-by-flow
detector. In the other, sequences of flows within a time window are considered
together to construct a window-based detector. These two methods will be called
\emph{flow 1-class SVM} and \emph{window 1-class SVM} method, respectively.

Finally, we also present a clustering method based on Adaptive Resonance Theory (ART)~\cite{Carpenter1987},
which is a machine learning technique originating in the biology field. 
This algorithm, named \emph{ART clustering}~\cite{art_tools}, partitions
network traffic into clusters based on the unique features of the network
flows. 

\begin{figure}
\centering
\includegraphics[width=8cm]{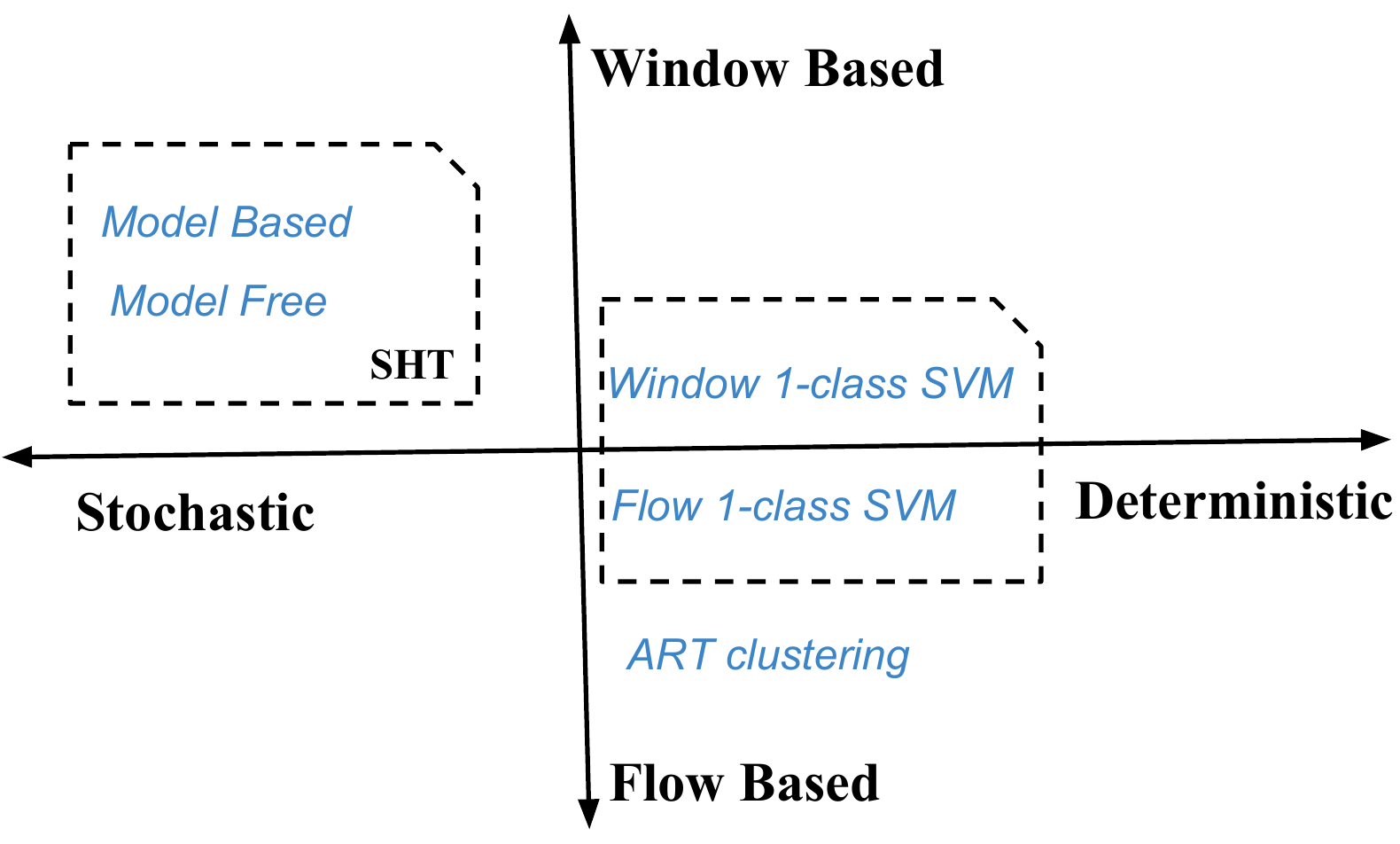}
\caption{Relationships among the five evaluated methods.\label{fig:relationship_methods}}
\vspace{-15pt}
\end{figure}

The relationships among these methods are depicted in
Figure~\ref{fig:relationship_methods}. The \emph{flow 1-class SVM} and the
\emph{ART clustering} method are flow-based and capable of identifying
individual network flows that are anomalous. By contrast, the remaining
methods are window-based, under which the flows are grouped into a window
based on their start time and only suspicious windows of time can be
identified as anomalous. \emph{Model-free} and \emph{model-based} methods are
stochastic and the rest methods are deterministic.

A challenging problem in the evaluation of anomaly detection methods is the
lack of test data with ground truth, due to the limited availability of such
data. The most widely used labeled dataset, DAPRA intrusion detection
dataset~\cite{Lippmann2000}, was collected 14 years ago. Since then, the
network condition has changed significantly. In order to address this problem,
we developed software to generate labeled data, including a flow-level anomaly
data generator (SADIT~\cite{sadit}) and a packet-level botnet attack data
generator (IMALSE~\cite{imalse}).  We evaluate all of our methodologies on a
simulated network and compare their performance under three flow-level
anomalies and one Distributed Denial of Service (DDoS) attack.

The rest of the paper is organized as follows. Section~\ref{sec:net_traff_rep}
describes the representation of network traffic data.
Section~\ref{sec:ano_det_methods} provides a mathematical description of the
methods used to identify anomalies.  Section~\ref{sec:net_sim} provides an
in-depth explanation of the simulated network and the anomalies.
Section~\ref{sec:results} presents the results of the five methods on the
simulated network data. Finally,
Section~\ref{sec:conclusion} provides concluding remarks.

\section{Network Traffic Representation}
\label{sec:net_traff_rep}
Let $\msS=\{\s^1,\dots,\s^{|\msS|}\}$ denote the collection of all packets on
the server which is monitored, where each element of $\msS$ is one packet. We
focus on host-based anomaly detection, in which case we only care about the
user IP address, namely the destination IP addresses for the outgoing packets
and the source IP addresses for the incoming packets. Denote the user IP
address in a packet $\s^i$ as $\x^i $, whose format will be discussed later.
The size of $\s^i$ is $b^i \in [0, \infty)$ in bytes and the start time of
transmission is $t^i \in [0, \infty)$ in seconds. Using this convention, the packet
$\s^i$ can be represented as $(\x^i, b^i, t^i)$ for all $i = 1,\ldots,|\msS|$.

Due to the vast number of packets, we consolidate this representation of
network traffic by grouping series of packets into flows. We compile a sequence
of packets $\s^1=(\x^1, b^1,t^1),\dots,\s^n=(\x^n,b^n,t^n)$ with
$t^1<\cdots<t^n$ into a \emph{flow} $\f=(\x, b, d_t, t)$ if $\x=\x^1=\cdots=\x^n$ and
$t^i-t^{i-1}<\delta_F$ for $i=2,\dots,n$ and some prescribed $\delta_F \in (0,
\infty)$. Here, the size $b$ is simply the sum of the sizes of the packets that
comprise the flow. The value $d_t=t^n-t^1$ denotes the flow duration. The
value $t=t^1$ denotes the start time of the first packet of the flow. In this
way, we can translate the large collection of traffic packets $\msS$ into a
relatively small collection of flows $\msF=\{\f^1,\dots,\f^{|\msF|}\}$.

In some applications in which large numbers of users frequently access the
server under surveillance, it may be infeasible to characterize network
behavior for each user. Different methods deal with this dilemma differently.

For both statistical and SVM methods, we first distill the
``user space'' into something more manageable while enabling us to characterize
network behavior of user groups instead of just individual users. For
simplicity of notation, we only consider IPv4 addresses. If $\x^{i} = (x_1^i,x_2^i,x_3^i,x_4^i)\in\{0,1,\dots,255\}^4 $ and
$\x^{j} = (x_1^j,x_2^j,x_3^j,x_4^j)\in\{0,1,\dots,255\}^4$ are two IPv4
addresses, the \emph{distance} between them is defined as:
$d(\x^i,\x^j)=\sum_{k=1,\dots,4}256^{4-k}|x_k^i-x_k^j|$.
This metric can be easily extended to IPv6 addresses if needed.
Suppose $\msX$ is the set of unique IP addresses in $\msF$.
We apply typical $K$-means clustering on $\msX$ \cite{hartigan1979, Lloyd82leastsquares}.
For each $\x\in\msX$, we thus obtain a cluster label $k(\x)$.
Suppose the cluster center for cluster $k$ is $\bar{\x}^k$, then the distance of
$\x$ to the corresponding cluster center is $d_a(\x)=d(\x, \bar{\x}^{k(\x)})$.
Using user clusters, we can produce our final representation of a flow as:
\begin{equation}
    \f=(k(\x),d_a(\x),b,d_t, t).
    \label{eq:flow_distill_def}
\end{equation}

For the \emph{ART clustering} method, distilling the user space beforehand is
not required. However, instead of using the IP address directly, we use a
compact representation.
Let $n_f(\x)$ be the number of flows transmitted between the user with IP
address $\x$ and the server. Define $d_b(\x) = d(\x, \x^*),\forall
\x\in\msX$, where $\x^*$ is the IP address of the server we are monitoring;
then the alternative flow representation we
use is:
\begin{equation}
    \f=(n_f(\x), d_b(\x),b,d_t, t).
    \label{eq:flowdef}
\end{equation}

\section{Anomaly Detection Methods}
\label{sec:ano_det_methods}

\subsection{Statistical Methods}
\label{subsec:statistical}
Let $h$ be the interval between the start points of two consecutive time windows and
$w_s$ be an appropriate window size; then the total number of windows is $n_w =
\left\lceil (t^{|\msF|} - t^1 - w_s) / h\right\rceil$. We say flow $\f^i =
(k(\x^i),d_a(\x^i), b^i, d_t^i, t^i)$ belongs to window $j$ if $\ t^1 + (j-1)h \leq
t^i < t^1 + (j-1)h + w_s,\ \forall j=1,\dots, n_w$. 

Let $\g^i = (k(\x^i),d_a(\x^i), b^i, d_t^i)$ be the flow attributes in $\f^i$
without the start time $t^i$ and $\msG_{j} = \{\g^1, \g^2, \dots,
\g^{|\msG_j|}\}$ be the flows in window $j$.  Let $\msG_{ref}$ be the set of
all flows used as reference.
The window-based methods will compare $\msG_j$ with $\msG_{ref}$ for all $j =
1, \dots n_w$.  Both statistical methods we will present in this section fall
into this category and can work in supervised as well as unsupervised modes. In
supervised mode, $\msG_j$ is generated by removing suspicious flows from a
small fragment of data through human inspection. In unsupervised mode, we
assume that the anomales are short-lived thus $\msG_j$ can be chosen as a
large set of nework traffic.

Since the approach introduced in what follows applies to all windows as
well as to nominal flows, we use $\msG=\{\g^1,\dots,\g^{|\msG|}\}$ to refer to 
$\msG_{ref}$ and $\msG_j, \forall j = 1, \dots n_w$.
Suppose the range of $d_a(\x^i),\forall i=1,\ldots,|\msG|$ is $[d_a^{min},
d_a^{max}]$. We can then define a discrete alphabet $\Sigma_{d_a} =
\left\{d_a^{min} + (m + 1/2)\times (d_a^{max} - d_a^{min}) /
|\Sigma_{d_a}|\right\}_{m=0,\ldots,|\Sigma_{d_a}|-1}$ for $d_a(\x^i)$, where
$|\Sigma_{d_a}|$ is called quantization level.  $\Sigma_{b}$ and $\Sigma_{d_t}$
can be defined similarly for $b^i$ and $d_t^i$. We then quantize $d_a(\x^i)$,
$b^i$ and $d_t^i$ in $\g^i$ to the
closest symbol in the discrete alphabet set $\Sigma_{d_a}$ and $\Sigma_{b}$ and
$\Sigma_{d_t}$, respectively. Suppose the total number of user clusters is
$K$. Then we can denote the quantized flow sequence
$\msG=\{\g^1,\dots,\g^{|\msG|}\}$ as
$\Sigma(\msG)=\{\bsigma(\g^1),\dots,\bsigma(\g^{|\msG|})\}$, where $\Sigma =
\{0, \dots, K-1\} \times \Sigma_{d_a} \times \Sigma_{b} \times \Sigma_{d_t} $
is the discrete alphabet for quantization
where each symbol in $\Sigma$ corresponds to a flow state.

\subsubsection{Model-free Method}
\label{subsubsec:modelfree}
In cases in which all flows emanating from the server under surveillance are
i.i.d., we construct the \emph{empirical measure} of flow sequence
$\msG=\{\g^1,\dots,\g^{|\msG|}\}$ as the frequency distribution vector 
\begin{equation}
    \msE^{\msG}(\brho)=\frac{1}{|\msG|}\sum_{i=1}^{|\msG|}\mathbf{1}\{\bsigma(\g^i)=\brho\},
\label{eq:iidempmeas}
\end{equation}
where $\mathbf{1}\{\cdot\}$ denotes the indicator function and $\bsigma(\g^i)$
denotes the flow state in $\Sigma$ that $\g^i$ gets mapped to. We will
denote the probability vector derived from the empirical measure of the form
in (\ref{eq:iidempmeas}) as
$\bmsE^{\msG}=\left\{\msE^{\msG}(\bsigma^1),\dots,\msE^{\msG}(\bsigma^{|\Sigma|})\right\}$.

Let $\bmu$ denote the probability vector calculated from the reference flows
$\msG_{ref}$. That is, $\mu(\sigma)$ is the reference marginal probability of
flow state $\sigma$. Using Sanov's
theorem~\cite{LockeWangPasTech,dembo2009large}, we construct a metric to
compare empirical measures of the form in (\ref{eq:iidempmeas}) to $\bmu$, thus
a metric of the ``normality'' of a sequence of flows. For every probability
vector $\bnu$ with support $\Sigma$, let
$H(\bnu|\bmu) =
\sum_{\bsigma\in\Sigma}\nu(\bsigma)\log\left(\nu(\bsigma)/\mu(\bsigma)\right)$
be the relative entropy of $\bnu$ with respect to $\bmu$.
Allowing $\eta=-\frac{1}{n}\log\epsilon$, where $\epsilon$
is a tolerable false alarm rate, then the \emph{model-free} anomaly detector is:
\begin{equation}
        \msI(\msG) = \mathbf{1}\{I_1(\bmsE^{\msG}) \geq \eta\},
    \label{eq:iiddectest}
\end{equation}
where $I_1(\bmsE^{\msG}) = H(\bmsE^{\msG}|\bmu)$.  It was shown in
\cite{Paschalidis2009} that (\ref{eq:iiddectest}) is asymptotically
Neyman-Pearson optimal.

\subsubsection{Model-based Method}
\label{subsubsec:modelbased}
As an alternative to the i.i.d.\ assumption on the sequence of flows under the 
\emph{model-free} method, we now turn to the case in which the sequence of
flows adheres to a first-order Markov chain. The notion of empirical measure
on the sequence $\msG=\{\g^1,\dots,\g^{|\msG|}\}$ must now be adapted to
consider subsequent pairs of flow states. We assume no knowledge of an initial
flow state $\bsigma(\g^1)$ and define the empirical measure on $\msG$, under
the Markovian assumption, as the frequency distribution on the possible flow
state transitions,
\begin{equation}
\msE_B^{\msG}(\bsigma^i,\bsigma^j)=\frac{1}{|\msG|}\sum_{l=2}^{|\msG|}\mathbf{1}\left\{\bsigma(\g^{l-1})=\bsigma^i,
\bsigma(\g^l)=\bsigma^j\right\},
\label{eq:markempmeas}
\end{equation}
where $\mathbf{1}\{\cdot\}$ denotes the indicator function and $\bsigma(\g^l)$
denotes the flow state in $\Sigma$ which $\g^l$ gets mapped to. We will denote
probability matrices formed by the empirical measure in (\ref{eq:markempmeas})
as
$\bmsE_B^{\msG}=\left\{\msE_B^{\msG}(\bsigma^i, \bsigma^j)\right\}_{i,j=1,\dots,|\Sigma|}$.

In the following, we will refer to matrices of the form
$\Q=\{q(\bsigma^i,\bsigma^j)\}_{i,j=1,\dots,|\Sigma|}$ as probability matrices
with support $\Sigma\times\Sigma$. By design, the empirical measures of the
form (\ref{eq:markempmeas}) are probability matrices with support
$\Sigma\times\Sigma$. Each probability matrix, under the Markovian assumption,
is associated with a transition probability matrix of the form
$\{q(\bsigma^j|\bsigma^i)\}_{i,j=1,\dots,|\Sigma|}$ where
$
q(\bsigma^j|\bsigma^i)=q(\bsigma^i,\bsigma^j) / q(\bsigma^i).
$
Here $q(\bsigma^i)=\sum_{j=1}^{|\Sigma|}q(\bsigma^i,\bsigma^j)$ denotes the
marginal probability of flow state $\bsigma^i$
in $\Q$. 

Let $\bPi=\{\pi(\bsigma^i,\bsigma^j)\}_{i,j=1,\dots,|\Sigma|}$ denote, under
the Markovian assumption, the true probability matrix of sequences of flows. As
in the i.i.d.\ case, we compute $\bPi$ via (\ref{eq:markempmeas}) from
$\msG_{ref}$. Following a similar procedure as in the i.i.d.\ case, we use an
analog of the Sanov's Theorem for the Markovian case, which appears
in~\cite{dembo2009large}, as the basis for our model-based stochastic anomaly
detector. 
For every shift invariant probability matrix $\Q$ with support $\Sigma\times\Sigma$, let
\begin{equation*}
H_B(\Q|\bPi)=\sum_{i,j=1}^{|\Sigma|}q(\bsigma^i,\bsigma^j)\log\frac{q(\bsigma^j|\bsigma^i)}{\pi(\bsigma^j|\bsigma^i)}.
\end{equation*}
be the relative entropy of $\Q$ with respect to $\bPi$.
Then in the \emph{model-based} method, the indicator of
anomaly for $\msG$ is:
\begin{equation}
\msI_B(\msG) = \mathbf{1}\{ I_2(\bmsE_B^{\msG}) \geq \eta \}
\label{eq:markdectest}
\end{equation}
where $I_2(\bmsE_B^{\msG}) = H_B(\bmsE_B^{\msG}|\bPi)$ and
$\eta=\frac{1}{n}\log\epsilon$ with $\epsilon$ be an allowable false alarm
rate. Again, the \emph{model-based} detector has been proved in
\cite{Paschalidis2009} to be asymptotically Neyman-Pearson optimal.

\subsection{1-class SVM}
\label{sec:1-class_SVM}
We turn now to deterministic methods based on the construction of a decision
boundary. 
We focus on one popular technique named 1-class
SVM~\cite{LockeWangPasTech, scholkopf}.
The premise behind 1-class SVM is to find a hyperplane that separates
the majority of the data $\msZ = \{\z^1, \dots, \z^{|\msZ|}\}$ from the
outliers by solving a Quadratic Programming (QP)
Problem~\cite{Paschalidis2009, scholkopf}. The hyperplane can be generalized to
a nonlinear boundary by mapping the inputs into high-dimensional spaces with a
kernel function $K(\cdot,\cdot)$~\cite{hastie2001elements}. There is a tunable parameter $\nu$
effectively tuning the number of outliers.

\subsubsection{Flow 1-class SVM}
\label{subsubsec:flowbyflow} 
We consider a set of flows $\msG=\{\g^1,\dots,\g^{|\msG|}\}$ that need to be
evaluated. According to (\ref{eq:flow_distill_def}), each flow has the format of
$\g=(k(\x),d_a(\x),b,d_t)$, which has already provided a rather compact
representation of network traffic. The only additional process required is to
remove the label of the cluster each user belongs to. The new data are:
$
\msZ=\{\z^1=(d_a(\x^1),b^1,d_t^1),\dots,\z^{|\msZ|}=(d_a(\x^{|\msZ|}),b^{|\msZ|},d_t^{|\msZ|}).
$
The reasoning for this is that, since we are measuring departures from nominal
users, the actual cluster a user belongs to is less important than the
distance between the user the cluster center. Besides, as a categorical
attribute, cluster labels make 1-class SVM method more unstable in practice.
Besides, we choose the radial basis function
$K(\u,\v)=\exp\left(-\gamma(\u-\v)^T(\u-\v)\right)$ as the kernel function~\cite{LockeWangPasTech}.

\subsubsection{Window 1-class SVM}
We combine the techniques described in Section~\ref{subsec:statistical} and the
1-class SVM into a window-based 1-class SVM method.  For each window $j$
with flows $\msG_j$, we can get the \emph{model-free} empirical measure
$\bmsE^{\msG_j}$ and the \emph{model-based} empirical measure
$\bmsE_B^{\msG_j}$. 
Let the feature vector for window $j$ be $Y^j=\left\{ \bmsE^{\msG_j},
\bmsE_B^{\msG_j}, |\msG_j| \right\}$.  Let $\msY=\{Y^1, \dots, Y^{|\msY|}\}$ be
a time series consisting of the features for all windows, then an 1-class SVM
can be used to evaluate $\msY$, resulting in a window-based anomaly detector.
Note that since the dimension of feature $Y^j$ is usually very large, it often helps to
apply Principal Component Analysis (PCA)~\cite{pearson1901liii} to reduce the
dimensionality first. 

\subsection{ART Clustering}
In this section, we present a clustering algorithm based on ART
theory~\cite{Carpenter1987} and apply it to network anomaly detection. The
algorithm first organizes inputs into clusters based on a customized distance metric.
Then, a dynamic learning approach is used to update clusters or to create new
clusters.

Assume a set of flows $\msF = \{\f^1, \ldots, \f^{|\msF|}\}$ with form in
(\ref{eq:flowdef}). Similar to the statistical methods, we define $\g^i
=(n_f(\x), d_b(\x),b^i,d_t^i)$  to be the attributes in $\f^i$
without the start time $t^i$ and $\msG$ is the counterpart of $\msF$. Suppose
$g^{ij}$ is the $j$th attribute of flow $\g^i$ for all $i=1,\ldots,|\msG|$ and
$j=1,2,3,4$. Defining $f_{min}(j, \msG)$ and $f_{max}(j, \msG)$ to be the
minimum and maximum of the set $\{ g^{ij}: \forall i=1,\ldots,|\msG|\}$, we can
normalize $\msG$ according to $\hat{g}^{ij} = \frac{g^{ij}-f_{min}(j,
\msG))}{f_{max}(j, \msG)-f_{min}(j, \msG)}$ for all $i=1,\ldots,|\msG|$, and
$j=1,2,3,4$. In this section we assume that the data in $\msG$ has already
been normalized.

Define the distance metric
\begin{equation}
    D(\p, \q) = \sum_{j=1}^m \left(\frac{p_j - q_j}{1-v_j}\right)^2
\end{equation}
for two $m$-dimensional vectors $\p=(p_1, \ldots, p_m)$ and $\q=(q_1, \ldots,
q_m)$, where $\v = (v_1, \ldots, v_m)$ is a set of parameters $v_j\in [0, 1)$
that controls the \emph{vigilance} in dimension $j$. 
Let $\mathcal{T}_k$ be the set of all flows in cluster $k$. Letting
$\c_k=(n_f^k, d^k , b^k, d_t^k) $ represent the center of cluster $k$ and
$c_k^j$ be its $j$ component.
Let $\msC$ be the set of all cluster centers. For every $\c \in \msC$ and a
prescribed $r$, 
\begin{equation}
    D(\g, \c) \leq r, \g\in\mathbb{R}^m
    \label{eq:ellipsoid}
\end{equation}
defines an ellipsoid in
$\mathbb{R}^m$. A higher vigilance in one dimension means the
ellipsoid is more shallow in this direction.

The \emph{ART clustering} algorithm is shown in Algorithm~\ref{alg:ARTC}.
Initially $\msC$ is empty. For each flow $\g^i \in \msG$, we calculate the set
$\msD$ which consists of all clusters whose ellipsoid defined
by~(\ref{eq:ellipsoid}) contains $\g^i$.  Suppose $E(\g, \c)$ is the Euclidean
distance between $\g$ and $\c$.  If $\msD$ is not empty,  $\g^i$ is assigned
to the cluster whose center has the smallest Euclidean distance with $\g^i$
and the corresponding cluster center is updated; otherwise a new cluster is
created. Suppose that flow $\g^i$ will be assigned to cluster $k$, let
$c_k^{j'}$ and $c_k^{j}$ be the $j$th component of the center of cluster $k$
before and after the assignment, then
\begin{equation} 
    c_{k}^{j'} = (p\times c_{k}^{j} + g^{ij})/(p+1),\forall j=1,\ldots,m,
    \label{eq:dyn_prop_cal}
\end{equation}
where $p$ is the number of flows in cluster $k$ before the assignment. Because
of the adaptive update (\ref{eq:dyn_prop_cal}), some assignments may become
unreasonable after update as some flows may become closer to other cluster
centers. As a result, the algorithm processes flows in $\msG$ again until an
equilibrium is reached.

\begin{algorithm}
\caption{ART clustering Algorithm}

\begin{algorithmic}
    \Require {Flow Data $\msG=\{\g^1, \ldots, \g^{|\msG|}\}$}
    \State {$\msC = \msC_{last} = \{\}$ }
    \While {$\msC \neq \msC_{last}$}
    \State {$\msC_{last} = \msC$}
    \For {$i=1 \to |\msG|$}
    \State {$\msD = \{\c \in \msC: D(\g^i, \c)) < r\}$}
    \If {$|\msD| = 0$}
        \State $\msC = \{\msC, \g^i\}$
    \Else
        \State $\c_o = \arg \min_{\c \in \msC } E(\g^i, \c)$
        \State $\msT_k = \{\msT_k, \g^i \}$, $k$ is the index of $\c_o$ in $\msC$.
        \State Recalculate cluster center of $\msT_k$ using (\ref{eq:dyn_prop_cal})
    \EndIf 
    \EndFor
    \EndWhile
\end{algorithmic}
\label{alg:ARTC}
\end{algorithm}
Once a stable equilibrium is reached,
small outlying clusters are identified as anomalous based on the rule 
\begin{equation}
    \msI_A(\msT_k) = \mathbf{1} \{|\msT_k| <
    \tau \times |\msG|/|\msC|  \}
     \label{eq:art_rule}
\end{equation}
where $\msI_A(\msT_k)$ is an indicator of anomaly for $\msT_k$, $\tau \in [0,1]$ is a prescribed detection threshold, $|\msC|$ and
$|\msG|$ are the total number of clusters and flows, respectively. 
$\tau$ determines how small a cluster must be to be considered as anomalous,
thus it influences the number of alarms. We will discuss the relationship of
$\tau$ and the false alarm rate further in Section~\ref{sec:results}.

\section{Network Simulation}
\label{sec:net_sim}
The lack of annotated data is a common problem in the network anomaly detection
community. As a result, we developed two open source software packages to
provide flow-level and packet-level validation datasets,
respectively. SADIT~\cite{sadit} is a software package containing all the
algorithms we described above. It also provides an annotated flow record
generator powered by the \emph{fs}~\cite{Sommers2011} simulator.
IMALSE~\cite{imalse} uses the NS3 simulator~\cite{henderson2008network} for the
network simulation and generates packet-level annotated data.  Simulation at the
packet-level takes more computation resources but can mimic certain attacks,
like botnet-based attacks, in a more realistic way.
We validate our algorithms with the help of these two software packages.
The packets generated by IMALSE, which is of \emph{pcap}
format~\cite{imalse}, are transformed into flow records first.  Then the
flows generated by SADIT and IMASLE are tested independently with each
algorithm. 



\begin{figure}[t]
    \centering
    \includegraphics[width=0.98\columnwidth]{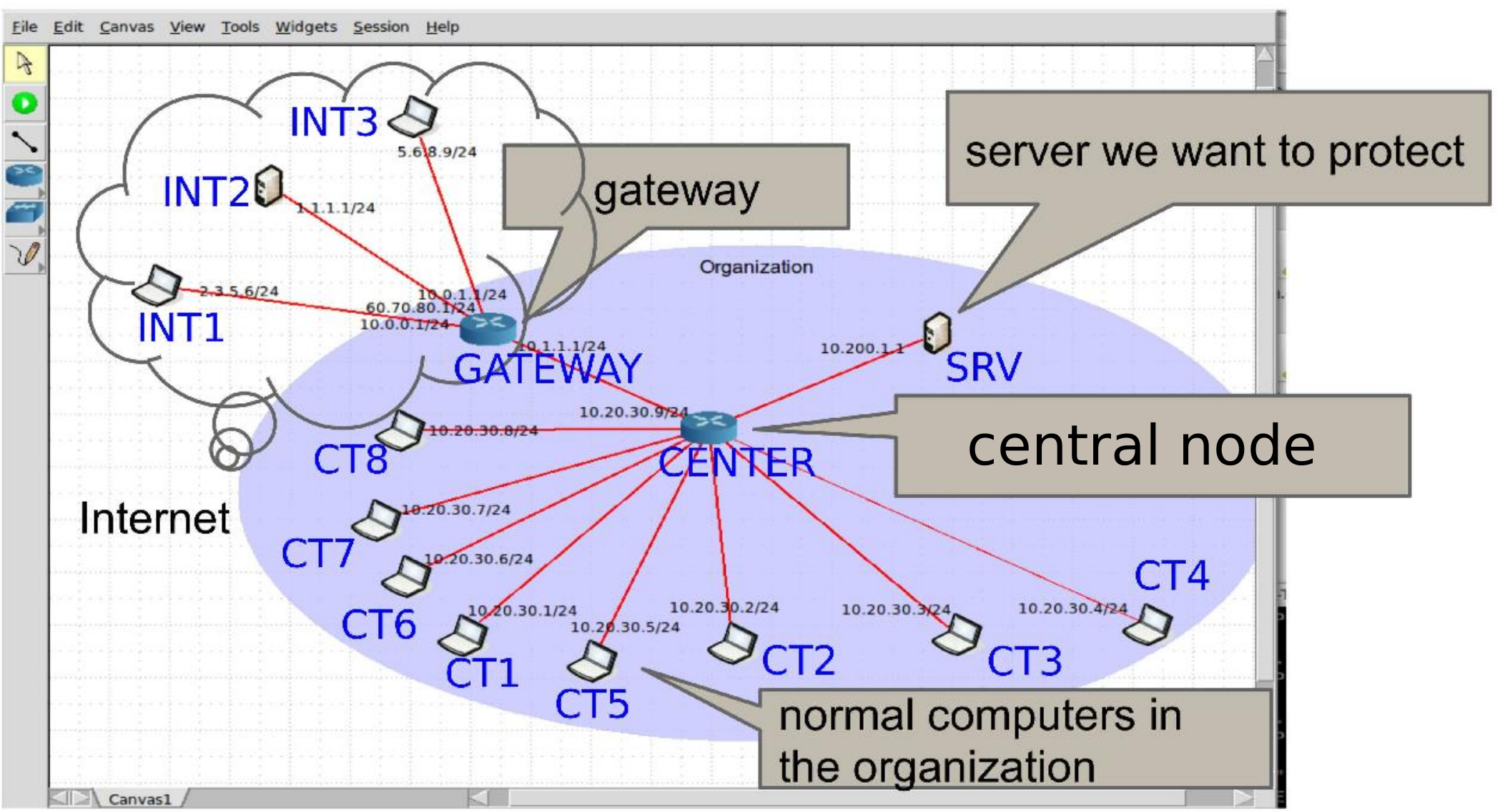}
    \vspace{-10pt}
    \caption{Simulation setting.}
    \vspace{-15pt}
    \label{fig:scene}
\end{figure}
The simulated network is partitioned into an internal network with a
hub and spoke topology that connects to the Internet via a gateway
(Fig.~\ref{fig:scene}). The internal network consists of 8 normal users
(\emph{CT1}-\emph{CT8}) and 1 server (\emph{SRV}) with some sensitive
information. We monitor the traffic on the server.

\subsection{Flow-level Anomalies}
First, we generate a dataset with flow-level anomalies.  The size and
the transmission of the nominal flows for user $i$ is assumed to follow a
Gaussian distribution $N(m_i, \sigma_i^2)$ and Poisson process with arrival
rate $\lambda_i$, respectively.  We investigate three most common types of
flow-level anomalies.

The first one mimics the scenario according to which a network intruder or
unauthorized user downloads restricted data. A previously unseen user who
has a large IP distance to the rest of the users starts transmission for a
short period. The second one is a user $i$ with suspicious flow size
distribution characterized by a mean $m_i^a$ higher than a typical value $m_i$.
Usually flows with substantially large flow size are associated with the
situation when some
users try to download large files from the server, which can happen when the
attacker tries to download the sensitive information packed into a
large file.
The last one is a user increasing its flow transmission rate to an unusual value
$\lambda_i^a$, which could be indicative of the user finding an important
directory on the server and downloading, repeatedly, sensitive files within
that directory.

\subsection{Packet-level Anomalies}
A second anomalous dataset is created using the tool
IMALSE~\cite{imalse}. The nominal traffic is generated using the
\emph{on-off} application in NS3~\cite{henderson2008network, imalse} in which the user
sends packets for $t_{on}$ seconds and the interval between two consecutive
transmission is $t_{off}$. The traffic is a Poisson process, which means the
\emph{on} time and \emph{off} times are exponentially distributed with parameter
$\lambda_{on}$ and $\lambda_{off}$, respectively.

We assume there is a botnet in the network. There is a botmaster controlling the bot
network and a Command and Control (C\&C) server issuing control commands
to the bots. In our simulation, both the botmaster and C\&C server are the
machine \emph{INT2} in the Internet, and \emph{CT1-CT5} in the internal
network have been infected as bots. We investigate a DDoS Ping flood attack in
which each bot sends a lot of ping packets to the server \emph{SRV} upon the
request of the C\&C server, aiming to exhaust the bandwidth of \emph{SRV}.
The attack is simulated at the packet-level and the data are then transformed into
flow records using techniques described in Section~\ref{sec:net_traff_rep}.
With appropriate $\delta_F$, the $t_{on}$ becomes the flow duration of nominal
flows and the $t_{off}$ determines the flow transmission rate of nominal
flows. The initiating stage of the attack is similar to the first case in the
previous section. During the attack, both the flow transmission rate and
the flow size of the bots may be affected. First, the flow transmission rate is
increased as the bots ping \emph{SRV} more frequently. Second, the ping packets
have different sizes from normal network traffic. Also, consecutive ping
packets may be combined together if they are sent over a short time interval.
The resulting flows may be very large in size if these combinations are common
or very small otherwise, depending on the attack pattern.


\section{Results}
\label{sec:results}
\subsection{Flow-level Anomalies}

\subsubsection{Atypical User}
\begin{figure}[t]
    \vspace{-15pt}
    \centering
    \includegraphics[width=0.4\textwidth, height=0.4\textwidth]{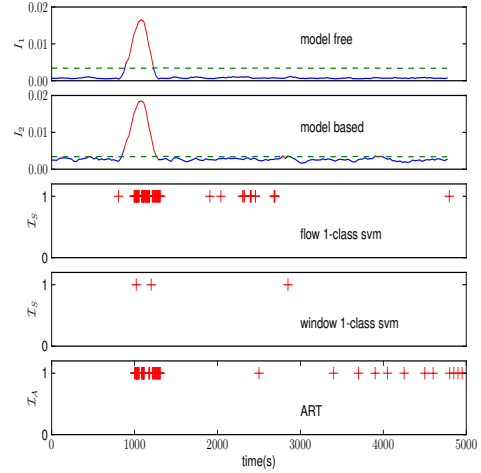}
    \vspace{-15pt}
    \caption{The results of five methods in the atypical user case.}
    \vspace{-15pt}
    \label{fig:atypical_user}
\end{figure}

Figure~\ref{fig:atypical_user} shows the response of all methods
described above when there is an atypical user trying to access the
server between 1000s and 1300s. For window-based methods, the interval
between the starting point of two consecutive time windows is $h=30s$
and the window size is chosen as $w_s=200s$, so there is overlap between
two consecutive time windows. We also distill the user space by using
$K$-Means clustering with 3 clusters. The quantization levels for flow
size, distance to cluster and flow duration are 3, 2, 1, respectively,
thus $|\Sigma|=18$.  The $x$-axis in all graphs corresponds to time (s)
and the total simulation time is 5000s. The first two graphs depict the
entropy metric in (\ref{eq:iiddectest}) and (\ref{eq:markdectest}) of the
\emph{model-free} and \emph{model-based} methods, respectively. For both
graphs, the green dashed line is the threshold when the false alarm rate
is $\epsilon=0.01$.  The interval during which the entropy curve is
above the threshold line (the red part) is the interval the method
reports as abnormal.  The $x$ coordinates of the red points with a `+'
marker correspond to the start point of the flow or the window the
method reports as abnormal. The parameter $\nu$ for the \emph{flow 1-class SVM}
and \emph{window 1-class SVM} is $0.002$ and $0.1$, respectively. The threshold
$\tau$ for \emph{ART clustering} is $0.05$.

We can observe from Figure~\ref{fig:atypical_user} that stochastic
methods, including our \emph{model-free} and \emph{model-based} methods,
tend to produce more stable results in the sense that they generate
fewer false alarms. At the same time, the \emph{flow 1-class SVM} and
\emph{ART clustering} methods, both of which are flow-based, can provide
higher identification resolution in the sense that they can identify the
suspicious flows, which is beyond the capabilities of the stochastic
methods. In the \emph{window 1-class SVM} method, we can tune the window
size to adjust the tradeoff of resolution and stability.  However, the
window size in the \emph{model-free} and \emph{model-based} methods has
to be reasonably large since the optimality of the decision rule
(\ref{eq:iiddectest}) and (\ref{eq:markdectest}) relies on the
assumption of a large flow number in each window.

\begin{figure}[h]
    \vspace{-10pt}
    \centering
    \includegraphics[width=0.4\textwidth, height=0.4\textwidth]{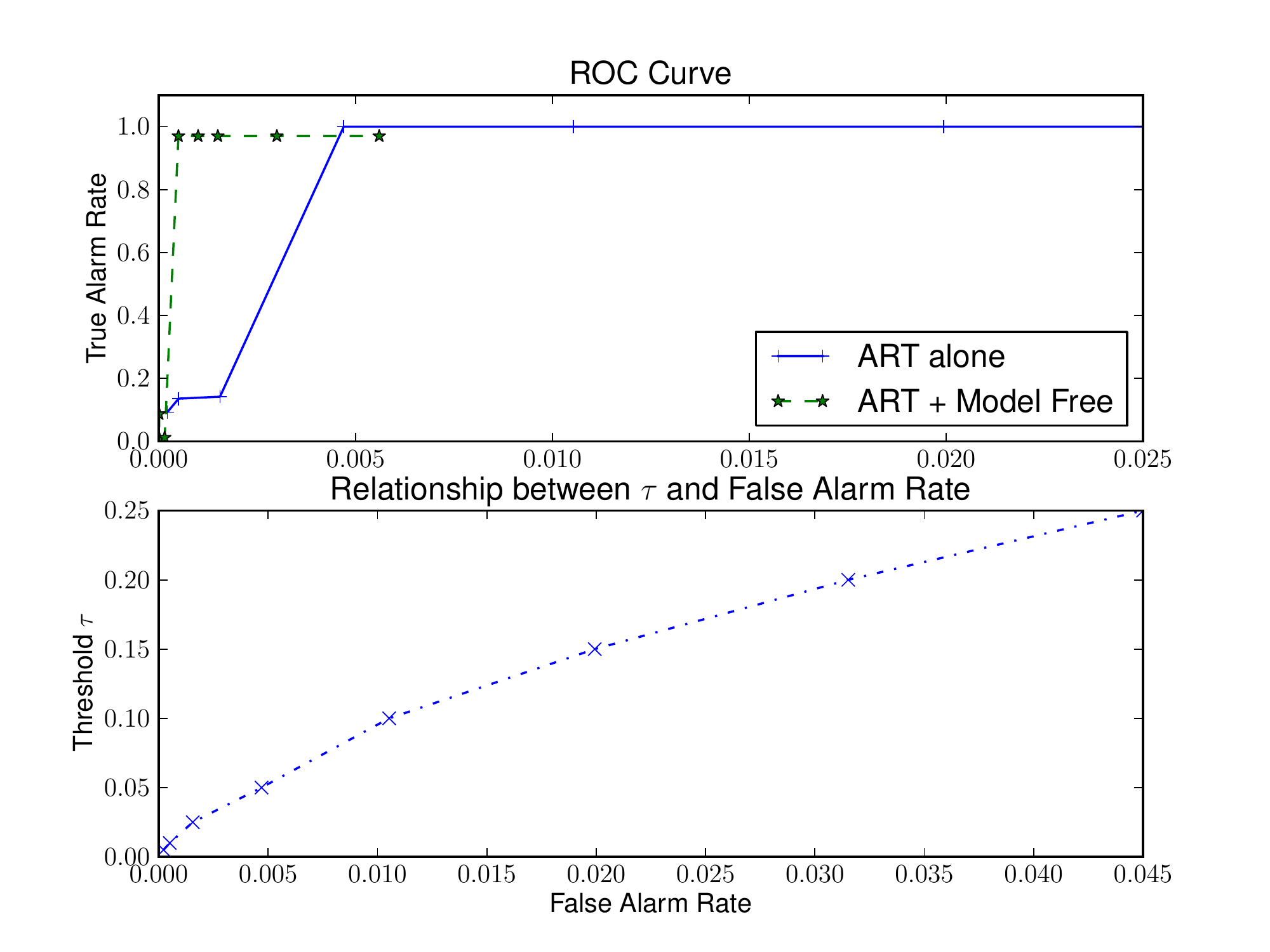}
    \vspace{-10pt}
    \caption{ROC curve and relationship of $\tau$ and the false alarm rate for ART.}
    \vspace{-5pt}
    \label{fig:art_specific}
\end{figure}

This observation indicates that these methods are complementary to each other.
One way to combine them is to use stochastic methods and window-based
deterministic methods to get a rough interval of an anomaly. Then, only the
flows that are both identified as suspicious by flow-based deterministic
methods and belong to the interval need to be further evaluated. The first
subfigure in Figure~\ref{fig:art_specific} shows the Receiver Operating
Characteristic (ROC) curve of the \emph{ART clustering} method, which is a flow-based method, and
the combination of the \emph{ART clustering} and the \emph{model-free} method.
The ROC curve has been substantially improved after combining the two methods.
The second subfigure in Figure~\ref{fig:art_specific} shows the relationship
between the threshold $\tau$ defined in (\ref{eq:art_rule}) and the false alarm
rate. The $x$-axis is the false alarm rate and $y$-axis corresponds to the threshold. As we
can see, the false alarm rate increases when the threshold increases and they are
almost linearly related to each other.

\subsubsection{Large File Download}
\begin{figure}[h!]
    \vspace{-7pt}
    \begin{center}
        \includegraphics[width=0.4\textwidth, height=0.3\textwidth]{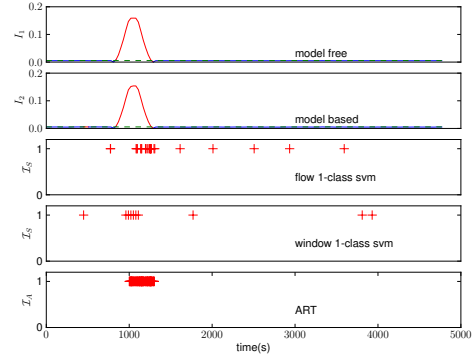}
    \end{center}
    \vspace{-10pt}
    \caption{The results of five methods in the large file download case.}
    \vspace{-10pt}
    \label{fig:large_file_download}
\end{figure}

Figure~\ref{fig:large_file_download} is the output of all methods in the
case where a user doubles its mean flow size between 1000s and 1300s.
Again, the first two graphs show the entropy curve and threshold line of the
\emph{model-free} and \emph{model-based} methods. The total simulation time is
5000s. The common window parameters $h$ and $w_s$ are the same as in the previous
case. The false alarm rate is $\epsilon=0.01$ for both \emph{model-free} and
\emph{model-based} methods. The parameter $\nu$ for \emph{flow 1-class SVM} and
\emph{window 1-class SVM} is 0.0015 and 0.1, respectively. $\tau=0.01$ for
\emph{ART clustering}.

\subsubsection{Large Access Rate}
\begin{figure}[h!]
    \begin{center}
        \includegraphics[width=0.4\textwidth, height=0.3\textwidth]{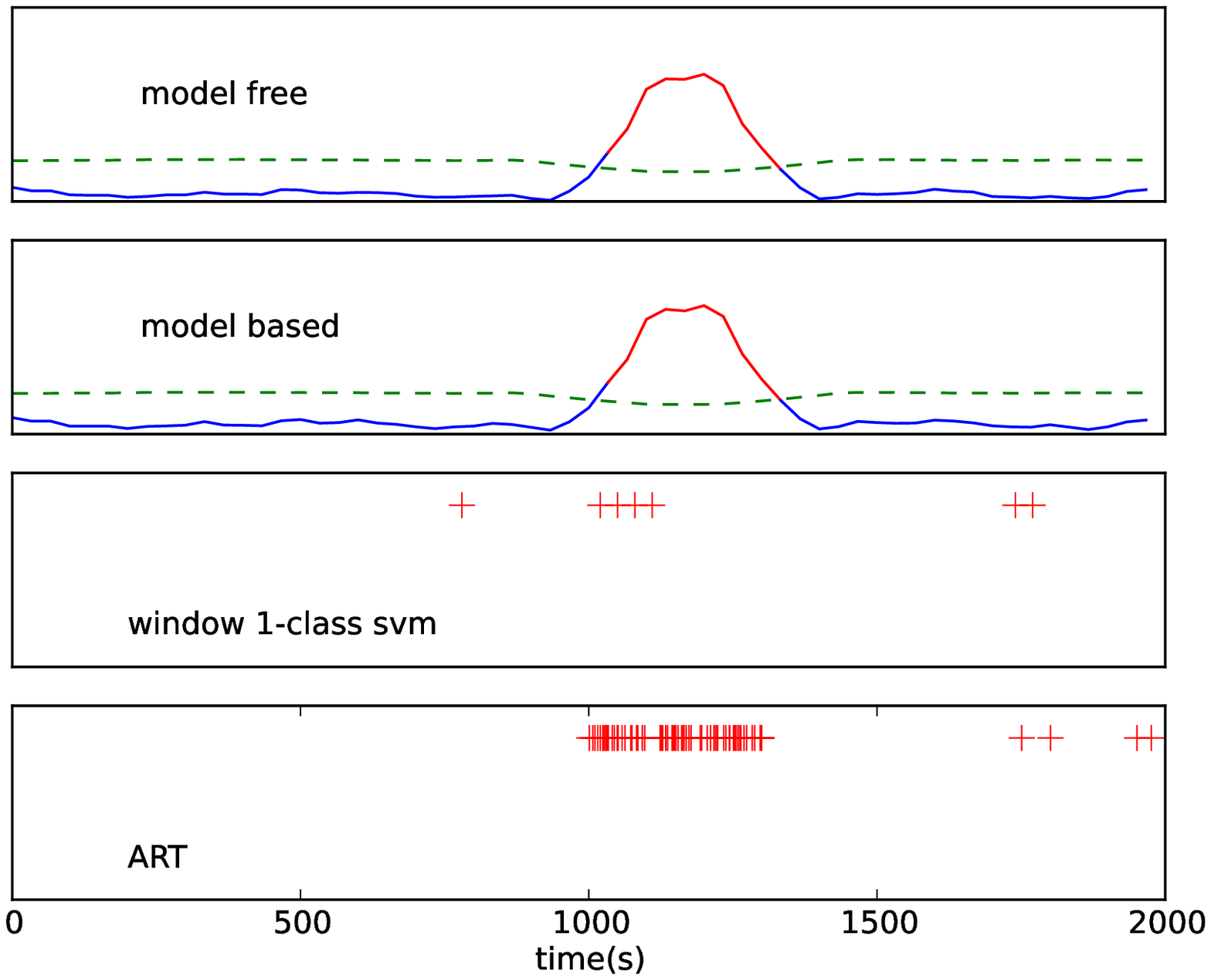}
    \end{center}
    \vspace{-10pt}
    \caption{The results of five methods in the large access rate case.}
    \vspace{-10pt}
    \label{fig:large_access_rate}
\end{figure}

Figure~\ref{fig:large_access_rate} shows the response of \emph{model-free},
\emph{model-based}, \emph{window 1-class SVM} and \emph{ART clustering} methods
when a user suspiciously increases its access rate to 6 times of its normal
value during 1000s and 1300s. The total simulation time is 2000s. The
parameters for the algorithms are the same in the atypical user case.

Note that \emph{flow 1-class SVM} cannot work for this type of anomaly since it
is purely temporal-based. The flow itself does not change but its frequency
does. There is no way to identify the frequency change by just observing the
individual flows with representation in (\ref{eq:flow_distill_def}). \emph{ART
clustering} works fairly well for this case because the attacker will have
larger $n_f(\x)$ as it transmits more flows. Interestingly, the \emph{model-based}
and \emph{model-free} methods can work very
well since the portion of traffic originating from the attacker changes, influencing
the empirical measure defined in (\ref{eq:iidempmeas}) and (\ref{eq:markempmeas}).
The two methods will not be effective in the very rare case when all users
increase their rate by the same ratio synchronously.

\subsection{DDoS Attack}
\begin{figure}[h]
    \begin{center}
        \includegraphics[width=0.4\textwidth]{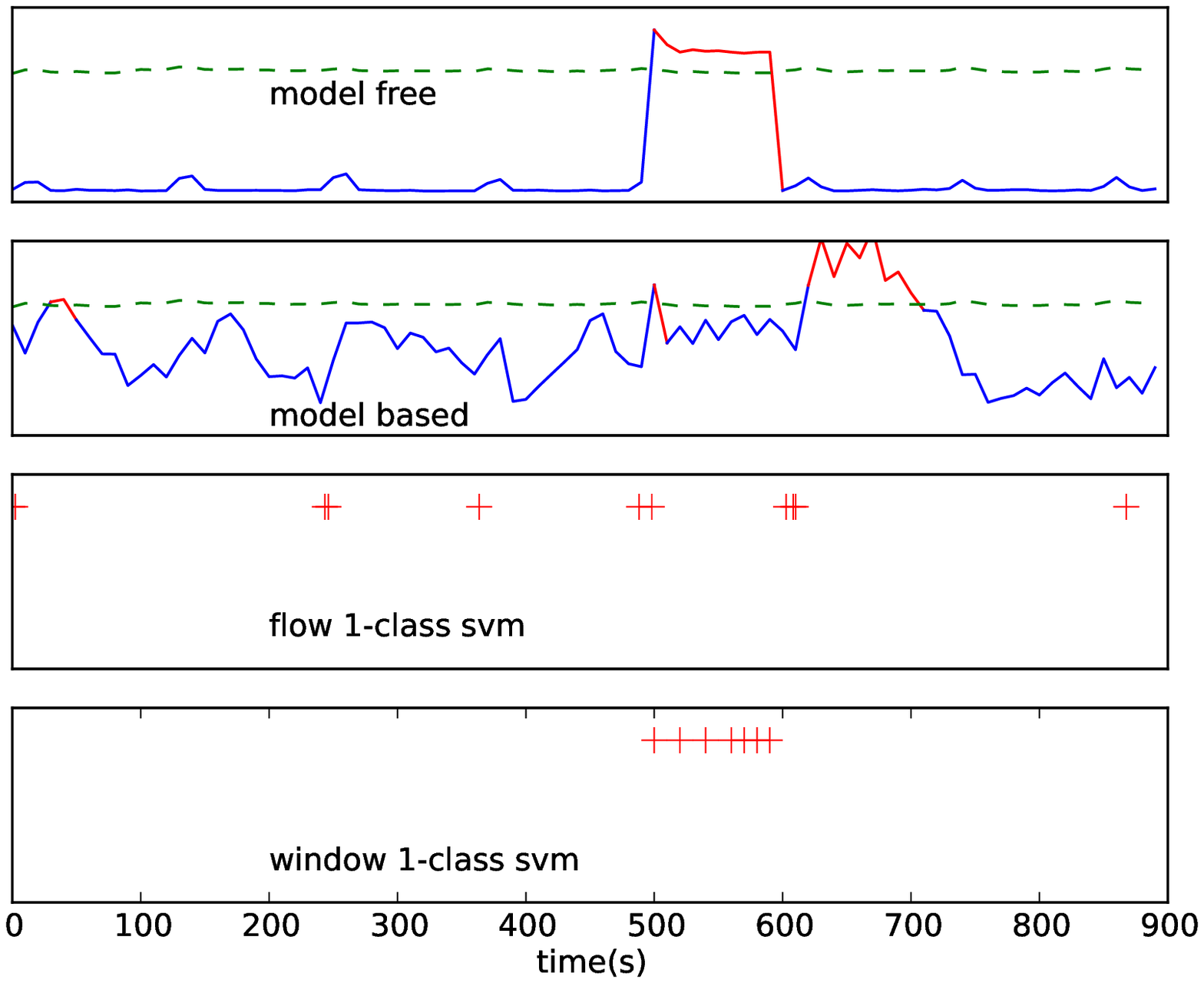}
    \end{center}
    \vspace{-10pt}
    \caption{The results for DDoS attack.}
    \vspace{-10pt}
    \label{fig:ddos_attack}
\end{figure}

Figure~\ref{fig:ddos_attack} shows the response of \emph{model-free},
\emph{model-based}, \emph{window 1-class SVM} and \emph{flow 1-class SVM}
methods when there is a DDoS attack targeting \emph{SRV} between 500s and 600s.
The total simulation time is 900s. For window-based methods, the interval
between consecutive time windows is $h=10s$ and the window size is $w_s=100s$. The
false alarm rate for the \emph{model-free} and \emph{model-based} method is
$\epsilon=0.01$ and $\nu =
0.05$ for \emph{window SVM}.

Since the nominal traffic in IMALSE is generated based on an i.i.d assumption,
it is hard for the \emph{model-based} method to capture a Markov
model. Yet, the \emph{model-based} method still detects the start
and the end of the attack, during which the transitional behavior changes the
most. \emph{Model-free} and \emph{window 1-class SVM} are more stable
while the \emph{flow 1-class SVM} method provides
higher resolution.

The \emph{ART clustering} method is also not suited to detect these type of attacks because the
unsupervised learning model is based on the assumption that malicious network
traffic represents a small percentage of total network traffic. A DDoS attack
generates a large number of packets and without some prior knowledge of good or
bad network traffic, the \emph{ART clustering} algorithm cannot distinguish between
the nominal and abnormal flows. It is also the reason for the relatively
unsatisfactory performance of the \emph{flow 1-class SVM} method. However,
\emph{window 1-class SVM} is not affected by this because despite the large
number of abnormal flows, the number of abnormal windows is still very small.


\section{Conclusion}
\label{sec:conclusion}
We presented five complementary approaches, based on SHT, SVM and clustering,
that cover the common techniques for host-based network anomaly detection. We
developed two open source software packages to provide flow-level and
packet-level validation datasets, respectively. With the help of these software
packages, we evaluated all methods on a simulated network mimicking typical
networks in organizations. We consider three flow-level anomalies and one
packet-level DDoS attack.

Through analyzing the results, we summarize the advantages and disadvantages of
each method. In general, deterministic and flow-based methods, such as
\emph{flow 1-class SVM} and \emph{ART clustering}, are more likely to have
unstable results with higher false alarm rates but they can identify abnormal
flows, namely they have better resolution. Stochastic and window-based methods,
such as our \emph{model-free} and \emph{model-based} methods, could yield more
stable results and detect temporal anomalies better, but they have relatively poor
resolution as they are not able to explicitly detect the anomalous network
flows. In addition, deterministic and window-based methods, like \emph{window
1-class SVM} offer parameters to adjust the tradeoff of resolution and
stability. This observation suggests that combining the results of all, instead
of just using one method, can yield better overall performance.


\bibliographystyle{abbrv}
\bibliography{cyber,web}

\end{document}